\documentclass{article}

\usepackage[preprint]{neurips_2026}
\usepackage[utf8]{inputenc}
\usepackage[T1]{fontenc}
\usepackage{hyperref}
\usepackage{amsmath}
\usepackage{graphicx}
\usepackage{url}
\usepackage{booktabs}
\usepackage{amsfonts}
\usepackage{nicefrac}
\usepackage{float}
\usepackage{microtype}
\usepackage{xcolor}
\usepackage{array}
\usepackage{colortbl}
\usepackage{multirow}

\title{CoCoT-EEG: Contrastive-Pretrained Multiscale Convolutional Transformer for EEG Decoding} 
\author{
  Gabriel Mahuas\thanks{Equal contribution.} \\
  Sigma Nova \\
  Paris, France \\
  \texttt{gabriel.mahuas@sigmanova.ai} \\
  \And
  Victoria Shevchenko\footnotemark[1] \\
  Sigma Nova \\
  Paris, France \\
  \texttt{victoria.shevchenko@sigmanova.ai} \\
  \And
  Ugo Tanielian\footnotemark[1] \\
  Sigma Nova \\
  Paris, France \\
  \texttt{ugo.tanielian@sigmanova.ai} \\
  \And
  Yassir Bendou \\
  Sigma Nova \\
  Paris, France \\
  \texttt{yassir.bendou@sigmanova.ai} \\
  \AND
  Richard Gao \\
  Goethe University Frankfurt \\
  Frankfurt, Germany \\
  \texttt{r.dg.gao@gmail.com} \\
}

\begin{document}

\maketitle

\begin{abstract}
  Self-supervised pretrained foundation models (FM) have shown early promise for non-invasive electroencephalogram (EEG) decoding applications. Many recent large-scale models converged on the approach of tokenizing raw EEG followed by masked reconstruction pretraining. However, this recipe has been shown to be suboptimal for data---like EEG---with high noise amplitude and information confined to limited dimensions such as narrow frequency bands. Building on this insight, we develop a novel contrastive-pretrained EEG model with multiscale temporal convolution input layers and Transformer encoder blocks (CoCoT). CoCoT matches or beats state-of-the-art reconstruction-pretrained EEG models on extensive benchmark decoding tasks with heterogeneous electrode configurations. Furthermore, CoCoT trained from scratch outperforms previous single-task decoding models and even rivals pretrained models, showcasing the architecture’s flexibility and data efficiency. Through systematic ablations, including model architecture and pretraining objective, we demonstrate the viability of contrastive learning for building EEG FMs while suggesting key architectural design considerations, prompting further investigations in alternative large-scale pretraining strategies.
\end{abstract}

\section{Introduction}
Electroencephalography (EEG) is pervasive in both research and clinical practice thanks to its moderate acquisition costs, non-invasive nature, and high temporal resolution.
Decoding from EEG data remains an important challenge which, when addressed, enables myriad applications such as brain-computer interface (BCI) and clinical diagnosis. 
Classical deep learning-based solutions use deep neural networks in a supervised setting while extracting information from the time series, e.g., using temporal convolutions as the inductive bias of choice \cite{Ermaganbet2023-wp, Lawhern2016-zo, Dung2021-hn, Ingolfsson2020-cq}.
While such solutions are generally effective, they are single-task models and often do not generalize well across settings \cite{Zhou2025-hl}.

Recently, large-scale self-supervised pretraining has proven successful at boosting cross-domain generalizability in language and computer vision. By leveraging vast amounts of unlabeled data, these models learn low-dimensional representations of their inputs that serve as generically useful features for downstream applications, often requiring little to no finetuning. Because of their general-purpose nature, such models have come to be known as foundation models (FMs) \cite{Bommasani2021-sx}.

This pretrained FM strategy has also been adopted for EEG decoding: recent models like LaBraM \cite{Jiang2024-qb}, CBraMod \cite{Wang2024-sh}, CSBrain \cite{Zhou2025-hl}, LUNA \cite{doner2025luna}, and REVE \cite{Ouahidi2025-gf} are pretrained on tens of thousands of hours of EEG data and perform well on a diverse set of downstream decoding tasks.
Inspired by vision and language models, these solutions have converged on a particular recipe of \textit{early patch tokenization} of the input data with \textit{reconstruction-based} pretraining objectives (e.g., masked autoencoding, MAE \cite{Kuruppu2025-nk}).
By further adapting this recipe to EEG data, such as discretizing into a codebook (via VQ-VAE \cite{Jiang2024-qb}), dedicated spatial/temporal attention \cite{Wang2024-sh, Zhou2025-hl}, and more sophisticated electrode position encoding strategies \cite{Ouahidi2025-gf, Zhou2025-hl} combined with increasingly massive pretraining datasets \cite{Ouahidi2025-gf}, pretrained EEG FMs have emerged as potential alternatives to tailored single-task models.

However, while this recipe of \textit{early-tokenization and reconstruction-based pretraining} has shown early promise, it ignores critical properties of EEG data: 
Specifically, EEG has high temporal resolution but low information density. 
Moreover, signal is often contained in narrow frequency bands and corrupted by noise sources such as muscle artifacts and sensor noise \cite{Lai2018-jb, Cohen2014-aw}. Thus, it's unclear if directly tokenizing and reconstructing EEG data is the most effective design for EEG FMs. 

In particular, recent work has shown both empirically and theoretically that reconstruction objectives bias models to focus on low-level details irrelevant for tasks relying on perceptual semantics \cite{Balestriero2024-bj}, and is especially counterproductive in high-noise regimes \cite{Van-Assel2025-nl}.
On the other hand, contrastive learning \cite{Radford2021-zb} has remained a viable strategy for pretraining language and vision models \cite{Radford2021-zb, Tschannen2025-zb, Jia2021-dr, Yu2022-wy}.
Furthermore, self-supervised contrastive objectives are more robust to noise, and are particularly effective when data augmentations can be designed based on prior knowledge of noise sources \cite{Van-Assel2025-nl}.
But while variants of contrastive pretraining has been employed in earlier EEG FMs like BENDR \cite{Kostas2021-hf} and BIOT \cite{Yang2023-hb}, those models similarly perform early patch tokenization on the signal, and do not leverage multi-scale convolutions to extract meaningful features from the low-SNR time series first.

\textbf{Main contributions:} Capitalizing on these insights and gaps, we develop a novel \textbf{Co}ntrastive-pretrained EEG model with multi-scale 1D \textbf{Co}nvolutions and \textbf{T}ransformer encoders (CoCoT-EEG):

1. Our \textit{convolution-first} architecture trained from scratch on individual datasets outperforms previous best single-task models, and for many tasks even surpasses large-scale pretrained EEG FMs.

2. \textit{Contrastive pretraining} further boosts performance on diverse downstream tasks, where CoCoT-EEG performs better than recent state-of-the-art reconstruction-pretrained models.

3. Through extensive ablation experiments, we isolate the effect of model architecture, pretraining objective, and pretraining dataset in addition to data augmentations for contrastive pretraining of EEG models, and analyze the generalizability of learned representations.

Overall, our results demonstrate the flexibility and data efficiency of the CoCoT-EEG base architecture, while offering insights on contrastive learning as an alternative pretraining strategy and design considerations for building large-scale, multi-task EEG decoding models.

\begin{figure*}
    \centering
    \includegraphics[width=1.0\textwidth]{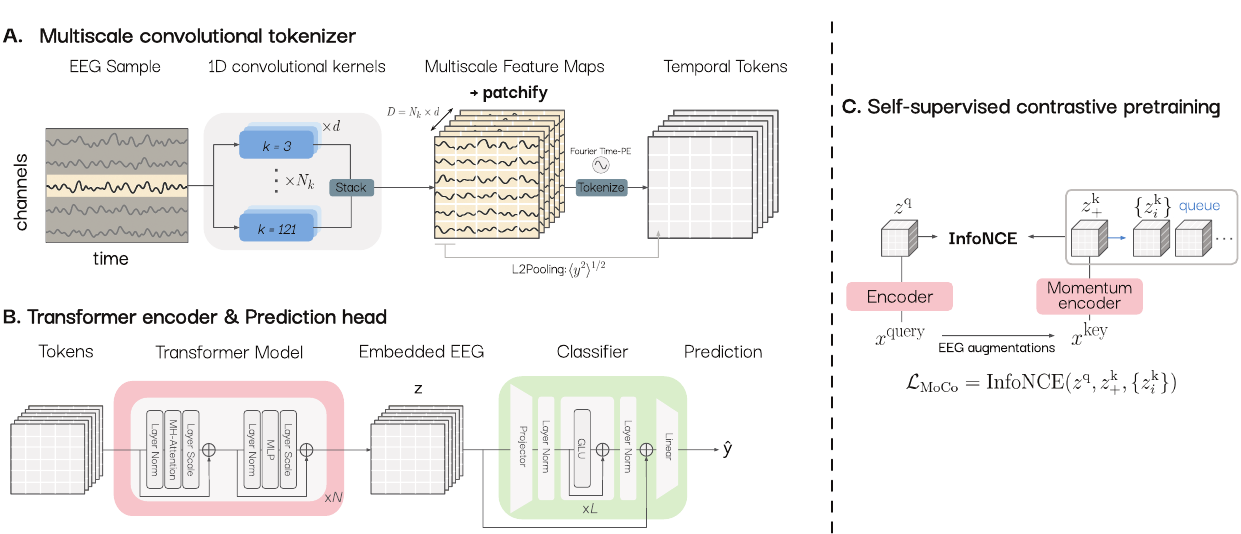}
    \caption{\textbf{CoCoT-EEG.} \textbf{A.} Raw EEG signals are processed through a bank of 1D convolutions with varying kernel sizes ($k=3$ to $121$). The resulting multiscale feature maps are patchified into temporal tokens via L2 pooling, followed by the addition of a Fourier positional encoding. \textbf{B.} Self-attention is applied to the tokens to produce EEG embeddings ($z$). A residual MLP head with two GLU blocks is used for the downstream classification or regression. \textbf{C.} For pretraining, a MoCo-style contrastive learning framework \cite{He2019-jw} is employed, where query $x^{query}$ and key $x^{key}$ samples are generated via EEG augmentations. The encoder and momentum encoder produce embeddings ($z^q$ and $z^k_+$), which are trained with InfoNCE \cite{van-den-Oord2018-nh} against a queue of negative samples $z^k_i$. GLU: gated linear unit; MH: multi-head, MoCo: momentum contrast; N: number of transformer blocks; PE: positional encoding.}
    \label{fig:model}
\end{figure*}

\section{Model architecture and training}
\label{sec:methods}

\subsection{Overview}
\label{sec:methods_overview}
We propose \textbf{CoCoT-EEG}, a novel convolution-Transformer hybrid architecture for EEG representation learning and decoding.
Given an EEG sample $X \in \mathbb{R}^{C \times T}$ (channels $C$, timepoints $T$), this model 
1) extracts multi-scale temporal features \textit{on the full sequence in raw signal space} with parallel 1D convolutional branches,
2) \textit{patchifies after embedding} through L2Pooling to produce tokens $Z \in \mathbb{R}^{C \times S \times D}$ with sequence length $S$ patches and embedding dimension $D$, for each channel $C$, and 3) \textit{encodes tokens} with a Transformer stack to produce per-token features. Here, given patch size $P$, the sequence length is expressed as $S=\lfloor T/P\rfloor$.

A central design choice is to \emph{embed before patchification}: unlike conventional tokenizers that segment the raw signal in patches prior to feature extraction, our convolution-first approach selectively filters out frequency-dependent noise \textit{and} preserves temporal continuity during early processing, especially long-timescale signals, effectively presenting later Transformer blocks with more information-dense features. Detailed descriptions of the model architecture are provided in Sec. \ref{sec:convolutions}-\ref{sec:transformer}. 

The model can be pretrained in a self-supervised manner using contrastive objectives (Sec. \ref{sec:moco}) and further finetuned with supervision labels, or similarly be trained from scratch end-to-end in a fully supervised setting. We ablate key model components and pretraining objective in Sec. \ref{sec:ablations}.

\subsection{Multiscale convolutional tokenizer}
\label{sec:convolutions}

\textbf{Multi-scale temporal branches.}
We use $d$ parallel temporal convolution branches with kernel sizes $\{k_i\}_{i=1}^{d}$, each operating on the {full} sequence:
$h_i = \mathrm{Conv1D}_{k_i}(X), \quad \text{with} \quad$ $h_i \in \mathbb{R}^{C \times D_i \times T}$
followed by group normalization and GELU nonlinearity.
Branches are of dimensions $D_i = D/d$ and are concatenated along the feature axis to yield multiscale features $H = [h_1,\ldots,h_d] \in \mathbb{R}^{C \times D \times T}$.

\textbf{Patchification in feature space.}
Patchification in feature space is performed by applying L$_2$ pooling with both kernel size and stride $P$ to produce
$Z \in \mathbb{R}^{C \times D \times S}$ whose entries are, for patch index $s\in\{0,\dots,S-1\}$ (a layer normalization to the token tensor is applied after that). 
\begin{equation}
\begin{aligned}
&Z[c,d,s]
=
\left(
\sum_{t=0}^{P-1}
\left| H[c,d,sP+t] \right|^{2}
\right)^{\frac{1}{2}},
\quad c\in\{1,\dots,C\},\ d\in\{1,\dots,D\}.
\end{aligned}
\end{equation}

\paragraph{Temporal Fourier Positional Encoding.}
\label{sec:temporal_fourier}
Positional information is injected using {temporal Fourier positional
encodings only}. Let $s \in \{0,\ldots,S-1\}$ denote the patch index along time. We build deterministic Fourier
features $\phi(s)\in\mathbb{R}^{D}$:
\begin{equation}
\phi_{2m}(s)=\sin\!\left(\frac{s}{\tau_m}\right), \quad
\phi_{2m+1}(s)=\cos\!\left(\frac{s}{\tau_m}\right),
\end{equation}
where $\{\tau_m\}$ are linearly spaced timescales. We then add the same temporal encoding to all channels: this provides temporal order information while keeping the model free of electrode geometry assumptions. See additional experiments with electrode position encoding schemes in Sec. \ref{sec:results_spatialPE}.

\subsection{Transformer Encoder \& Prediction head}
\label{sec:transformer}
\paragraph{Backbone.} The backbone encoder is a stack of $L$ transformer layers operating on $Z\in\mathbb{R}^{C\times S\times D}$. 
We apply multi-head self-attention over all channel--time tokens. Given token features $Z$, we flatten them into a sequence $\tilde{Z}$ of length $N=C\cdot S$. Each transformer layer applies MHSA followed by a position-wise feed-forward network (FFN). We use a pre-layer normalization design with residual connections, stochastic depth (DropPath), and learnable LayerScale parameters initialized to $10^{-4}$. The FFN employs a gated SwiGLU activation. We then discard the projector and use the backbone as a general EEG encoder. For downstream tasks, we compute token features $Z\in\mathbb{R}^{C\times S\times D}$ and apply a task-specific head.

\paragraph{Residual MLP head for supervised finetuning.}
For downstream classification or regression, we flatten the final-layer token features before applying an MLP head composed of $L_h$ residual GLU blocks.
The flattened features are first projected to a hidden dimension $d_{\text{proj}}$ with GELU and dropout; residual GLU blocks with scaled residual factor $\alpha$ are then applied, followed by a skip connection from the input and a final linear prediction layer after layer normalization.

\subsection{Self-Supervised Contrastive Pretraining with MoCo}
\label{sec:moco}
\textbf{Objective.} We pretrain the encoder in a {self-supervised} manner using Momentum Contrast (MoCo) \cite{He2019-jw}.
Given an unlabeled EEG segment $X$, we sample two stochastic augmentations to produce two views
$X^q = t_1(X)$ and $X^k = t_2(X)$.
The query encoder $f_\theta$ and key encoder $f_{\bar{\theta}}$ (updated through an exponential moving average of the query encoder, with momentum $m=0.999$) share the same architecture and we pool the token features into a single vector representation through attention pooling before passing it through a projection head $g(\cdot)$:
\begin{equation}
\begin{aligned}
&q = g(\mathrm{pool}(f_\theta(x^q))), \ &k = g(\mathrm{pool}(f_{\bar{\theta}}(x^k))).
\end{aligned}
\end{equation}
The projected vectors are then $\ell_2$-normalized before computing the contrastive loss. We maintain a queue of negative keys $\{k^-_i\}_{i=1}^{K}$ of length $K=65536$ and optimize the InfoNCE loss:
\begin{equation}
\mathcal{L}_{\text{MoCo}} =
-\log \frac{\exp(q^\top k^+ / \tau)}{\exp(q^\top k^+/\tau) + \sum_{i=1}^{K} \exp(q^\top k_i^- / \tau)},
\end{equation}
where $\tau$ is the temperature (set to $\tau=0.07$) and $k^+$ is the positive key corresponding to $q$.

\textbf{Augmentations.}
We construct the two views of each EEG sample by applying random compositions of EEG-specific augmentations. These augmentations mimic common acquisition artifacts and noise observed in EEG recordings. In particular, they encourage invariance to corrupted or missing electrodes, sensor noise, temporal occlusions, and moderate spectral perturbations. Our augmentation pipeline includes channel dropout, additive Gaussian noise, time masking, random per-channel scaling and offset, bandstop filtering, and phase jitter. See Appendix \ref{app:augmentations} for details.

\textbf{Pooling and projector.}
After the Transformer encoder, we implement a simplified attention pooling over the flattened token set. Given a flattened token matrix $U \in \mathbb{R}^{(CS) \times D}$, we compute attention weights by projecting each token onto a learnable vector $w \in \mathbb{R}^D$ (with a learnable scalar bias $b$) and applying a softmax across tokens; the pooled representation is the resulting weighted sum of tokens. We then pass this representation through a SimCLR-style two-layer MLP projector \cite{Chen2020-an} consisting of two linear layers with a ReLU activation and BatchNorm after the first layer, and no activation or normalization after the output layer.

\subsection{Data}

\paragraph{Pretraining datasets.}
For our main results, we pretrain on the Temple University EEG Corpus (TUEG) \cite{Obeid2016-az}, a clinical EEG dataset comprising over 16,000 sessions from more than 10,800 subjects and totaling approximately 27,000 hours of recordings, which has been a standard pretraining dataset for EEG FMs. We also perform an ablation study on the pretraining dataset by using the Healthy Brain Network (HBN) \cite{Shirazi2024-ym} dataset, which differ drastically from TUEG and comprises approximately 1,500 hours of 128-channel EEG recordings from 2,600 children and adolescents aged 5 to 21. We used the following tasks from HBN release 1-11: contrast change detection, resting state, surround suppression, and a movie watching session. Both datasets were downsampled to 100 Hz.

\paragraph{Downstream task datasets}
\label{downstream_data}
We evaluate CoCoT-EEG on 10 standard EEG foundation model benchmark tasks spanning: motor imagery classification (\texttt{BCI-IV}~\cite{tangermann2012review}, \texttt{PHYSIO}~\cite{Goldberger2000-ao}, \texttt{SHU}~\cite{Ma2022-hr}), emotion recognition (\texttt{FACED}~\cite{chen2023large}, \texttt{SEED-V}~\cite{liu2021comparing}),
seizure and abnormality detection (\texttt{CHB-MIT}~\cite{shoeb2009application}, \texttt{TUAB}~\cite{Obeid2016-az}), sleep stage classification (\texttt{HMC}~\cite{Alvarez-Estevez2021-df}, \texttt{ISRUC}~\cite{Khalighi2016-lr}), and vigilance estimation regression (\texttt{SEED-VIG}~\cite{Zheng2017-tu}).
These datasets range from 4 to 64 channels and have between 10K to 409K samples.
Appendix~\ref{app:downstream} and Table~\ref{tab:downstream-tasks} provide an overview and detailed information about the benchmark datasets. We note that \texttt{TUAB} is a subset of TUEG. Similar to the baseline models \cite{Wang2024-sh, Ouahidi2025-gf, Zhou2025-hl} we compare against that use \texttt{TUEG} for pretraining without excluding \texttt{TUAB}, we do not use the class labels in any way during pretraining. Nevertheless, we evaluate with an independent pretraining dataset in HBN (Appendix \ref{tab:fullwithhbn}).

\paragraph{Data preprocessing}
\label{sec:preproc}
Note that downstream datasets were split into train, validation and test following the same process implemented in \cite{Wang2024-sh}. One important note is that we deviate from their preprocessing strategy by using a robust centering and normalization strategy with median and median absolute deviation (MAD) per channel for robustness to outliers and artifacts common in clinical EEG recordings: $x_{\text{norm}} = (x - \text{median}_{\text{ch}}) / (\text{MAD}_{\text{ch}} + \epsilon)$, where $\epsilon = 10^{-6}$.

\section{Experiments}
We evaluate our model on a diverse suite of EEG datasets and decoding tasks spanning multiple configurations and application domains (Sec.~\ref{downstream_data}, Appendix \ref{app:downstream}), while comparing to recent state-of-the-art (SOTA) models. In this section, we first describe the baseline models we compare against (Sec \ref{sec:baselines}). Then, we show that our model is performant in both the ``from-scratch'' and pretrained (Sec. \ref{sec:results_pretrained}) settings compared to single-task and foundation models, respectively. Finally, we perform extensive ablation experiments to explore the impact of pretraining objective.

\subsection{Baseline models}\label{sec:baselines}
We compare CoCoT-EEG trained from scratch against competitive single-task model performances reported in \cite{Zhou2025-hl}, including ContraWR \cite{Yang2023-hb}, CNN-Transformer \cite{Peh2022-bm}, and Spatiotemporal-Transformer \cite{Song2021-gh}, which achieved best-of-class performance on one or more tasks. In the pretrained setting, we compare against recent SOTA EEG foundation models, including BIOT \cite{Yang2023-lg}, LaBraM \cite{Jiang2024-qb}, CBraMod \cite{Wang2024-sh}, REVE \cite{Ouahidi2025-gf}, and CSBrain \cite{Zhou2025-hl}. Note that CBraMod and CSBrain are both pretrained on curated subsets of TUEG (roughly 9,000 hours) while REVE is pretrained on a custom dataset of around 60,000 hours compiled from various public repositories (see Appendix \ref{app:pretrainHP} and Appendix \ref{app:computing_resources} for details). 

\subsection{From scratch and pretrained model performance}\label{sec:results_pretrained}

\paragraph{From scratch.} We first evaluated CoCoT-EEG trained from scratch on each downstream task independently. CoCoT-EEG strongly outperforms other single-task models on most tasks (Table~\ref{tab:expanded_from_scratch_baselines}), including those with similar design patterns such as CNN-Transformer and Spatiotemporal-Transformer. Even when compared against foundation models, our architecture without any pretraining achieved competitive performance across most tasks (Table~\ref{tab:main_results_extended}, \texttt{SCRATCH}).
\begin{table*}[h]
\caption{Performance of CoCoT-EEG trained from scratch, or pretrained (PT) with MoCo on TUEG and finetuned. Balanced accuracy (in \%) is reported for all tasks except SEED-VIG (Pearson corr.). Baseline performances are taken from published values in \cite{Ouahidi2025-gf} \cite{Zhou2025-hl}. Mean ± 95\% CI is reported for all models except REVE (mean ± std). \textbf{Bold} denotes SOTA mean. FT: end-to-end fine-tuning. $\star$ denotes runs using generalization finetuning techniques such as Mixup or Euclidean-Alignment \cite{HeWu-euclidean}}
\label{tab:main_results}
\vskip 0.15in
\begin{center}
\begin{small}
\begin{sc}
\setlength{\aboverulesep}{0pt}
\setlength{\belowrulesep}{0pt}
\setlength{\tabcolsep}{2.8pt}
\newcolumntype{g}{>{\columncolor[gray]{0.9}}c}
\begin{tabular}{lgggggclc}
\toprule
 & \multicolumn{5}{g}{Baselines} & & \multicolumn{2}{c}{CoCoT-EEG (OURS)} \\
\cmidrule{2-6} \cmidrule{8-9}
\rowcolor{white}
Dataset & \cellcolor[gray]{0.9}\small BIOT & \cellcolor[gray]{0.9}\small LaBraM & \cellcolor[gray]{0.9}\small CBraMod & \cellcolor[gray]{0.9}\small REVE & \cellcolor[gray]{0.9}\small CSBrain & & \small Scratch & \small TUEG PT \\
\midrule
BCI-IV   & 47.48 {\tiny $\pm$ 0.93} & 47.48 {\tiny $\pm$ 0.85} & 51.38 {\tiny $\pm$ 0.66} & \textbf{63.96$^\star$} {\tiny $\pm$ 0.95} & 56.57 {\tiny $\pm$ 0.71} & & 54.40 {\tiny $\pm$ 2.99} & 60.55 {\tiny $\pm$ 1.40} \\
CHB-MIT  & 70.68 {\tiny $\pm$ 4.57} & 70.75 {\tiny $\pm$ 3.58} & 73.98 {\tiny $\pm$ 2.84} & --- & 72.62 {\tiny $\pm$ 1.15} & & 60.55 {\tiny $\pm$ 0.79} & \textbf{82.35} {\tiny $\pm$ 0.61} \\
FACED    & 51.18 {\tiny $\pm$ 1.18} & 52.73 {\tiny $\pm$ 1.07} & 55.09 {\tiny $\pm$ 0.89} & 56.46$^\star$ {\tiny $\pm$ 1.64} & 57.52 {\tiny $\pm$ 0.42} & & 55.56 {\tiny $\pm$ 0.73} & \textbf{57.73} {\tiny $\pm$ 2.04} \\
HMC      & 68.62 {\tiny $\pm$ 0.41} & 72.77 {\tiny $\pm$ 1.01} & 72.69 {\tiny $\pm$ 0.41} & --- & 73.45 {\tiny $\pm$ 0.47} & & 73.69 {\tiny $\pm$ 0.21} & \textbf{76.13} {\tiny $\pm$ 0.11} \\
ISRUC    & 75.27 {\tiny $\pm$ 1.21} & 76.33 {\tiny $\pm$ 1.02} & 78.65 {\tiny $\pm$ 1.1} & --- & 79.25 {\tiny $\pm$ 0.30} & & 74.62 {\tiny $\pm$ 0.32} & \textbf{79.97} {\tiny $\pm$ 0.34} \\
PHYSIO   & 61.53 {\tiny $\pm$ 1.53} & 61.73 {\tiny $\pm$ 1.22} & 64.17 {\tiny $\pm$ 0.91} & 64.80$^\star$ {\tiny $\pm$ 1.50} & 63.04 {\tiny $\pm$ 0.90} & & 62.44 {\tiny $\pm$ 1.31} & \textbf{65.31} {\tiny $\pm$ 1.42} \\
SEED-V   & 38.37 {\tiny $\pm$ 1.87} & 39.76 {\tiny $\pm$ 1.38} & 40.91 {\tiny $\pm$ 0.97} & --- & 41.97 {\tiny $\pm$ 0.33} & & 41.41 {\tiny $\pm$ 0.44} & \textbf{42.13} {\tiny $\pm$ 0.36} \\
SHU      & 61.79 {\tiny $\pm$ 1.83} & 61.66 {\tiny $\pm$ 1.92} & 63.70 {\tiny $\pm$ 1.51} & --- & 64.17 {\tiny $\pm$ 0.37} & & 62.25 {\tiny $\pm$ 1.65} & \textbf{64.74} {\tiny $\pm$ 0.87} \\
TUAB     & 79.59 {\tiny $\pm$ 0.57} & 81.40 {\tiny $\pm$ 0.19} & 82.49 {\tiny $\pm$ 0.25} & \textbf{83.15}$^\star$ {\tiny $\pm$ 0.15} & 81.72 {\tiny $\pm$ 0.43} & & 79.17 {\tiny $\pm$ 0.14} & 82.78 {\tiny $\pm$ 0.25} \\
SEEDVIG & 61.14 {\tiny $\pm$ 1.69} & 63.47 {\tiny $\pm$ 1.35} & 55.02 {\tiny $\pm$ 1.15} & --- & 63.14 {\tiny $\pm$ 3.56} & & 60.37 {\tiny $\pm$ 2.04} & \textbf{64.43} {\tiny $\pm$ 1.12} \\
\bottomrule
\end{tabular}
\end{sc}
\end{small}
\end{center}
\vskip -0.1in
\end{table*}

\paragraph{Pretrained.} We next assessed the impact of contrastive pretraining for CoCoT-EEG. 
First, following standards in the literature, we pretrained using TUEG (clinical and around 27,000 hours of EEG data), which yielded significant improvements on all downstream tasks. Compared to EEG FMs that used the same pretraining dataset, CoCoT-EEG outperforms CBraMod and CSBrain on all tasks. We also note that CoCoT-EEG performs significantly better than BIOT, a previous EEG FM following BERT-style contrastive pretraining but with early tokenization.

Even when compared to REVE, which is pretrained on 60,000 hours of diverse EEG data and finetuned with additional regularization strategies, including Mixup \cite{Zhang2017-bs} and Euclidean Alignment \cite{HeWu-euclidean}, CoCoT-EEG pretrained on TUEG and finetuned with ``vanilla'' strategies perform comparably, showcasing its data efficiency. CoCoT-EEG finetuned with Mixup further improves performance, achieving $83.50 {\tiny \pm 0.15}$ on TUAB (+0.72\%, surpassing REVE), $63.20 {\tiny \pm 0.55}$ for FACED (+5.47\%), and $68.7 {\tiny \pm 1.19}$ for SEED-VIG (up +4.27\%). Overall, \textbf{CoCoT-EEG sets the new SOTA on 8/10 tasks} with no additional finetuning strategies, and \textbf{9/10} including TUAB with Mixup.

\paragraph{Varying pretraining dataset.}
We pretrained CoCoT-EEG on HBN instead of TUEG to evaluate whether the benefits of pretraining occur across substantially different dataset configurations. 
Compared to TUEG, HBN has a very different distribution of subjects in the healthy development context (5-21 years old), with less recording hours in total (around 1,500 hours from the subset of tasks we chose), and a different electrode configuration of 128 channels. 
Even so, HBN pretraining still provided generalizable benefits when finetuned on downstream tasks (Appendix~\ref{tab:fullwithhbn}).

\subsection{Ablation experiments and analysis}\label{sec:results_analysis}
\label{sec:ablations}
Seeing the benefits of our multi-scale convolution and Transformer architecture with contrastive pretraining, we systematically ablate each component of CoCoT-EEG to analyze their contributions:

\subsubsection{Pretraining objective ablation: contrastive vs. MAE}
We first compare contrastive (MoCo) and reconstruction (MAE) pretraining objectives while holding the CoCoT-EEG architecture fixed (Table~\ref{tab:pretrained_ablation}, left most). For the MAE experiments, we adopt the  strategy and preprocessing pipeline of CBraMod~\cite{Wang2024-sh}; implementation details are provided in Appendix~\ref{app:mae}. MoCo is superior on the majority of downstream tasks, with the largest gains on BCI-IV ($+2.9$), FACED ($+7.9$), and SEED-VIG ($+5.3$).

Interestingly, the only exception is CHB-MIT, where MAE outperforms MoCo by $2.68$ points. This pattern is consistent with theoretical analysis showing that joint-embedding (contrastive) objectives should be preferred over reconstruction ones when the data contains high-magnitude but \textit{irrelevant} features \cite{Van-Assel2025-nl}. In CHB-MIT, however, the high-variance components likely coincide with the diagnostic signal, i.e., high-amplitude seizure events. This suggests that the choice of pretraining objective interacts non-trivially with the downstream task: contrastive pretraining may be preferred for general EEG classification, but reconstruction-based objectives retain an advantage for tasks with high SNR.

\begin{table*}[h]
\caption{Pretraining objective and architecture ablations. Comparing CoCoT-EEG's backbone with all kernels with MoCo (base) vs. MAE pretraining, as well as against Convolution-first and Patch-first variants with truncated convolution kernels. Pearson correlation for SEED-VIG, balanced accuracy (\%) for the rest. Values are mean $\pm$ 95\% CI. \textbf{Bold} denotes the best result per row.}
\label{tab:pretrained_ablation}
\vskip 0.15in
\begin{center}
\begin{small}
\begin{sc}
\setlength{\aboverulesep}{0pt}
\setlength{\belowrulesep}{0pt}
\setlength{\tabcolsep}{3pt}
\begin{tabular}{lccccc}
\toprule
& \multicolumn{2}{c}{All Kernels} & \multicolumn{3}{c}{Truncated Kernels} \\
\cmidrule(lr){2-3} \cmidrule(lr){4-6}
Dataset & MoCo & MAE & Conv-1st MoCo & Patch-1st MoCo & Patch-1st MAE \\
\midrule
BCI-IV   & \textbf{60.55} {\tiny $\pm$ 1.40} & 57.62 {\tiny $\pm$ 7.02} & 59.16 {\tiny $\pm$ 0.80} & 47.55 {\tiny $\pm$ 1.20} & 46.70 {\tiny $\pm$ 5.57} \\
FACED    & \textbf{57.73} {\tiny $\pm$ 2.04} & 49.82 {\tiny $\pm$ 4.97} & 55.58 {\tiny $\pm$ 1.13} & 50.98 {\tiny $\pm$ 1.05} & 47.14 {\tiny $\pm$ 2.86} \\
PHYSIO   & \textbf{65.31} {\tiny $\pm$ 1.42} & 63.26 {\tiny $\pm$ 1.87} & 64.24 {\tiny $\pm$ 1.10} & 62.66 {\tiny $\pm$ 1.20} & 59.01 {\tiny $\pm$ 2.06} \\
SEED-V   & \textbf{42.13} {\tiny $\pm$ 0.36} & 40.29 {\tiny $\pm$ 0.93} & 41.19 {\tiny $\pm$ 0.64} & 41.45 {\tiny $\pm$ 0.67} & 39.83 {\tiny $\pm$ 1.41} \\
CHB-MIT & 82.35 {\tiny $\pm$ 0.61} & \textbf{85.03} {\tiny $\pm$ 0.66} & 79.74 {\tiny $\pm$ 0.74} & 81.29 {\tiny $\pm$ 1.80} & 79.61 {\tiny $\pm$ 1.84} \\
SHU      & \textbf{64.74} {\tiny $\pm$ 0.87} & 61.92 {\tiny $\pm$ 1.99} & 64.61 {\tiny $\pm$ 0.72} & 63.56 {\tiny $\pm$ 1.26} & 61.28 {\tiny $\pm$ 0.57} \\
SEED-VIG & \textbf{64.43} {\tiny $\pm$ 1.12} & 59.13 {\tiny $\pm$ 3.55} & 64.09 {\tiny $\pm$ 3.11} & 64.48 {\tiny $\pm$ 1.68} & 62.71 {\tiny $\pm$ 2.42} \\
\bottomrule
\end{tabular}
\end{sc}
\end{small}
\end{center}
\vskip -0.1in
\end{table*}

\subsubsection{Model architecture ablation}\label{sec:results_arch}

\paragraph{Convolution-first, Patch-first, and Truncated kernels.}
To isolate the contribution of the CoCoT-EEG architecture from that of the pretraining objective, we evaluate two tokenization variants---a convolutional tokenizer (Conv-first) and a non-overlapping patch tokenizer (Patch-first)---both with truncated convolutional kernels, under matched MoCo and MAE pretraining (Table~\ref{tab:pretrained_ablation}).
Two patterns emerge consistently. First, CoCoT-EEG outperforms both truncated variants under MoCo pretraining, indicating that its full kernel set captures information that the truncated variants discard.
Second, when comparing the truncated architectures directly, Conv-first consistently outperforms Patch-first under both pretraining and supervised training from scratch.
Under MoCo pretraining, Conv-first performs better on 4/7 downstream tasks, often substantially (e.g., +11.6 on BCI-IV).
The same trend also holds in the fully supervised setting (Appendix \ref{app:trunc_scratch}), where Conv-first achieves stronger performance on 5/7 tasks.
Even with a reduced kernel bank, the convolutional tokenizer's locality bias captures the local time-frequency structure of EEG more effectively than fixed-size patches.

\paragraph{Spatial positional encoding ablation.} \label{sec:results_spatialPE}
CoCoT-EEG intentionally uses only temporal positional encoding and does not explicitly encode electrode topology. This design choice makes the model agnostic to montage configuration and avoids imposing fixed spatial assumptions across datasets with heterogeneous channel layouts. However, EEG signals exhibit strong spatial organization, and recent EEG FMs have proposed explicit electrode-aware positional encoding schemes \cite{Ouahidi2025-gf}. 

To evaluate whether incorporating electrode geometry improves representation learning, we augmented CoCoT-EEG with a 3D spatial positional encoding (3DPE, see Appendix~\ref{app:spatialPE}) inspired by \cite{Ouahidi2025-gf}, derived from electrode coordinates, and compared it against the default topology-agnostic design. We additionally evaluated a control setting in which electrode positions were randomly shuffled across channels \textit{once} before finetuning.
\begin{table}[h]
\caption{
Spatial positional encoding ablation. We benchmark a topology-aware CoCoT-EEG variant using a 3D spatial positional encoding (\textbf{3DPE}) and a control condition where electrode positions are randomly shuffled once before finetuning (\textbf{3DPE + Shuffled Pos.}). Results are reported as balanced accuracy (\%) except for SEED-VIG (Pearson correlation $\times$ 100). Values in bold exceed the performance of the topology-agnostic CoCoT-EEG baseline reported in Table~\ref{tab:main_results}.}
\label{tab:sptialPE_ablation}
\begin{center}
\begin{small}
\begin{sc}
\setlength{\tabcolsep}{3pt}
\begin{tabular}{lccc}
\toprule
Dataset 
& 3DPE + Shuffled Pos. 
& 3DPE \\
\midrule
BCI-IV 
& $57.23 \pm 3.01$ 
& $57.14 \pm 2.62$ \\

FACED 
& $57.15 \pm 1.41$ 
& $\mathbf{57.83 \pm 1.63}$ \\

PHYSIO 
& $64.95 \pm 1.44$ 
& $\mathbf{66.07 \pm 0.42}$ \\

SEED-V 
& $41.75 \pm 0.91$ 
& $\mathbf{42.21 \pm 0.57}$ \\

SHU 
& $64.72 \pm 2.04$ 
& $\mathbf{65.14 \pm 0.50}$ \\

SEED-VIG 
& $62.89 \pm 5.09$ 
& $62.92 \pm 3.53$ \\
\bottomrule
\end{tabular}
\end{sc}
\end{small}
\end{center}
\vskip -0.1in
\end{table}
We observe that explicit spatial encoding does not consistently improve downstream performance (Table~\ref{tab:sptialPE_ablation}). While modest gains are seen on some tasks (e.g., PhysioNet MI and SHU), performance remains unchanged or degrades on others. Furthermore, electrode positions shuffling does not consistently degrade performance, suggesting that the spatial positional encoding contributes only marginally to the learned representations in the current setting. 

\subsubsection{Pretraining augmentation ablation}\label{sec:results_augs}

To understand which data augmentations contribute most to the learned representations, we pretrained small-scale versions of our architecture using both a full augmentation pipeline and two complementary ablation settings: \emph{leave-one-out} (LOO), where a single augmentation is removed from the full pipeline, and \emph{leave-one-in} (LOI), where pretraining is performed using only a single augmentation.

\begin{table*}[h]
\caption{Effect of individual augmentations across datasets (balanced accuracy, \%).
``Leave-one-out'' reports the difference between the full
augmentation setup vs. \textit{without that augmentation}.
``Leave-one-in'' reports the difference between from scratch and pretraining with \textit{only that augmentation}. Positive values indicate improved performance. 
\textbf{Bold} denotes the best \textit{per column}.}
\label{tab:augmentation_ablation_all_datasets}
\begin{center}
\begin{sc}
\setlength{\tabcolsep}{1.5pt}
\begin{tabular}{lcccccc}
\toprule
& \multicolumn{3}{c}{Leave-one-out (Full $-$ Abl)}
& \multicolumn{3}{c}{Leave-one-in (Aug $-$ Scratch)} \\
\cmidrule(lr){2-4}
\cmidrule(lr){5-7}
Augs.
& SHU & Physio & FACED
& SHU & Physio & FACED \\
\midrule
Time Masking
& $\mathbf{+0.56}$ {\tiny $\pm 1.01$} & $\mathbf{+2.68}$ {\tiny $\pm 0.79$} & $\mathbf{+1.64}$ {\tiny $\pm 1.10$}
& $+0.07$ {\tiny $\pm 0.99$} & $\mathbf{+1.47}$ {\tiny $\pm 0.92$} & $-1.35$ {\tiny $\pm 1.55$} \\
Gauss.\ Noise
& $-0.19$ {\tiny $\pm 0.65$} & $+0.64$ {\tiny $\pm 0.70$} & $-0.08$ {\tiny $\pm 1.03$}
& $+0.11$ {\tiny $\pm 0.77$} & $+0.19$ {\tiny $\pm 0.67$} & $-0.57$ {\tiny $\pm 1.17$} \\
Ch. Dropout
& $-1.14$ {\tiny $\pm 1.26$} & $+0.20$ {\tiny $\pm 0.94$} & $+0.13$ {\tiny $\pm 0.91$}
& $+0.73$ {\tiny $\pm 1.50$} & $+0.81$ {\tiny $\pm 1.19$} & $-0.93$ {\tiny $\pm 1.10$} \\
Scale/Offset
& $-0.21$ {\tiny $\pm 1.52$} & $-0.11$ {\tiny $\pm 0.89$} & $+0.51$ {\tiny $\pm 1.50$}
& $\mathbf{+1.09}$ {\tiny $\pm 0.43$} & $+0.68$ {\tiny $\pm 0.75$} & $-0.61$ {\tiny $\pm 1.50$} \\
Band Stop
& $-0.66$ {\tiny $\pm 1.48$} & $-0.55$ {\tiny $\pm 1.07$} & $-0.68$ {\tiny $\pm 1.58$}
& $+0.01$ {\tiny $\pm 1.89$} & $-0.75$ {\tiny $\pm 1.72$} & $-1.37$ {\tiny $\pm 1.04$} \\
Phase Jitter
& $-0.12$ {\tiny $\pm 1.77$} & $+0.01$ {\tiny $\pm 0.34$} & $+0.24$ {\tiny $\pm 1.27$}
& $+1.01$ {\tiny $\pm 1.12$} & $+0.74$ {\tiny $\pm 0.94$} & $\mathbf{-0.55}$ {\tiny $\pm 1.57$} \\
\bottomrule
\end{tabular}
\end{sc}
\end{center}
\vskip -0.1in
\end{table*}
Overall, augmentation effects are strongly dataset-dependent (Table~\ref{tab:augmentation_ablation_all_datasets}), with Time Masking emerging as the most consistently beneficial augmentation, particularly on PhysioNet MI where its removal causes the largest degradation. Most other leave-one-out ablations produce small or no effects. In the leave-one-in setting, no single augmentation consistently recovers the benefit of the full pipeline: modest gains are observed on SHU and PhysioNet MI, whereas performance generally decreases on FACED. Taken together, these findings suggest that the gains from augmentation are driven more by augmentation diversity and complementary invariances than by any individual transformation alone.

\subsubsection{Linear Probing and learned embeddings}\label{sec:results_lp}
Table~\ref{tab:probing_vs_finetuning_delta} compares linear probing (LP) and full end-to-end finetuning (FT) across three pretraining setups. The LP$\rightarrow$FT gap ($\Delta$) is small on tasks that align well with the pretraining distribution, indicating that the pretrained representations are already linearly separable. As tasks drift further from the pretraining objective, the gap widens substantially.
SEED-VIG is particularly informative: it is a regression task rather than classification, so the linear head is solving a fundamentally different problem than the contrastive or reconstructive pretraining objective optimized for.

\begin{table*}[h]
\caption{Linear Probing (LP) vs.\ end-to-end finetuning (FT) across pretraining setups (Pearson correlation for SEED-VIG, balanced accuracy in \% for the rest, mean $\pm$ 95\% CI). $\Delta$ denotes gain from LP to FT. \textbf{Bold} denotes the best FT result per row.}
\label{tab:probing_vs_finetuning_delta}
\begin{center}
\begin{small}
\begin{sc}
\setlength{\tabcolsep}{2.5pt}
\begin{tabular}{lccccccccc}
\toprule
 & \multicolumn{3}{c}{TUEG MoCo} & \multicolumn{3}{c}{HBN MoCo} & \multicolumn{3}{c}{TUEG MAE} \\
\cmidrule(lr){2-4} \cmidrule(lr){5-7} \cmidrule(lr){8-10}
Dataset & LP & FT & $\Delta$ & LP & FT & $\Delta$ & LP & FT & $\Delta$ \\
\midrule
SHU      & 63.15 & \textbf{64.74} & $+1.59$  & 63.09 & 64.07 & $+0.98$  & 60.73 & 61.92 & $+1.19$ \\
PHYSIO   & 60.09 & \textbf{65.31} & $+5.22$  & 58.11 & 64.64 & $+6.53$  & 54.94 & 63.26 & $+8.32$ \\
BCI-IV   & 52.56 & \textbf{60.55} & $+7.99$  & 54.32 & 56.08 & $+1.76$  & 50.74 & {57.62} & $+6.88$ \\
SEED-V   & 36.55 & \textbf{42.13} & $+5.58$ & 32.78 & 40.74 & $+7.96$  & 32.30 & 40.29 & $+7.99$ \\
FACED    & 41.44 & \textbf{57.73} & $+16.29$ & 47.18 & 56.15 & $+8.97$  & 25.69 & 49.82 & $+24.13$ \\
SEED-VIG & 33.61 & \textbf{64.43} & $+30.82$ & 23.91 & 61.66 & $+37.75$ & 29.45 & 59.13 & $+29.68$ \\
\bottomrule
\end{tabular}
\end{sc}
\end{small}
\end{center}
\vskip -0.1in
\end{table*}
Additionally, the relative ordering of pretraining setups is also consistent across LP and FT, where MoCo results in representations that are also beneficial for Linear Probes (e.g., TUEG MoCo vs. TUEG MAE). These findings show that contrastive embeddings are also of high quality with minimal finetuning, while underscoring that the utility of pretraining varies across tasks: motor imagery and simpler emotion tasks benefit from direct transfer while fine-grained affective computing requires extensive finetuning to adapt representations to task-specific demands. 

\section{Discussion}
\subsection{Summary}
We introduce CoCoT-EEG, a contrastive-pretrained multiscale convolutional Transformer that challenges the prevailing reconstruction-based pretraining paradigm for EEG foundation models. Our architecture embeds the raw signal through parallel multi-scale 1D convolutions and tokenizes through L2 pooling, preserving temporal continuity and filtering noise before Transformer encoding. We report three main findings: First, the CoCoT architecture is data-efficient. Trained from scratch, it exceeds previous single-task models across benchmark tasks while approaching several pretrained FMs.
Second, contrastive pretraining provides consistent---and often substantial---performance gains compared to training from scratch. Third, pretrained CoCoT matches or surpasses recent reconstruction-pretrained SOTA baselines (CBraMod, REVE, CSBrain) across diverse downstream tasks. With extensive ablation experiments, we validate contrastive learning as a viable alternative to masked reconstruction for EEG foundation models. These results suggest that the field's convergence on masked autoencoding as the pretraining objective of choice may be premature, especially for high-noise, low-SNR signals like EEG.

\subsection{Related works and our contributions}

\paragraph{Self-Supervised EEG Foundation Models.}
Recent large-scale EEG FMs have predominantly been pretrained with reconstruction objectives \cite{Jiang2024-qb, Wang2024-sh, Ouahidi2025-gf}: One of the first to massively scale data and model capacity, LaBraM \cite{Jiang2024-qb} introduced vector-quantized neural spectrum prediction for masked EEG modeling, pretraining on $\sim$2,500 hours across 20 datasets. CBraMod \cite{Wang2024-sh} proposed a criss-cross Transformer that separately models spatiotemporal dependencies through parallel spatial and temporal attention, achieving strong generalization across 10 BCI tasks. REVE \cite{Ouahidi2025-gf} further scaled up to 60k hours of EEG recordings while leveraging 4D positional encoding to align heterogeneous datasets. Lastly, similar to CoCoT-EEG, CSBrain \cite{Zhou2025-hl} employs cross-scale feature extraction, but is MAE-pretrained and applies convolutions \textit{after patchification and only within short windows}, limiting the potential advantage of convolutions in extracting long-timescale information \textit{across} windows.

\paragraph{Theoretical Foundations of Contrastive Learning.}
Our work builds on theoretical insights from the joint embedding literature. Contrastive losses asymptotically optimize two key properties \cite{Wang2020-pj}: alignment, which ensures that semantically similar samples are mapped to nearby representations, and uniformity, which encourages features to be evenly distributed to preserve maximal information.  Furthermore, recent theoretical and empirical work suggests that joint-embedding methods are more robust than reconstruction-based objectives in regimes dominated by high-magnitude nuisance variability \cite{Van-Assel2025-nl}, a setting particularly relevant for EEG signals. Motivated by these observations, CoCoT-EEG combines multiscale convolutions that first extract features at the raw signal level with contrastive pretraining to learn noise-robust EEG representations.

\paragraph{Contrastive Learning for EEG modeling.}
Previous single-task models have, for example, explored relative positioning and temporal shuffling tasks for sleep staging \cite{Banville2020-pd}. 
Early contrastive self-supervised EEG FMs include BENDR \cite{Kostas2021-hf}, which adapts wav2vec2.0's framework \cite{Baevski2020-bs} to learn transferable representations.
Meanwhile, BIOT \cite{Yang2023-lg} follows the BERT recipe by first breaking the raw signal into small, fixed-size patches, then tokenizing and embedding their Fourier spectra while using a contrastive objective for pretraining.
More recently, hybrid objectives combining masked reconstruction with contrastive learning have been explored: CoMET \cite{li2025cometcontrastivemaskedbrainfoundation}, for instance, combines masked EEG modeling with contrastive objectives, while LEAD \cite{wang2026leadeegfoundationmodel} introduces a gated temporal-spatial Transformer specifically for EEG-based Alzheimer's disease diagnosis using medical contrastive pretraining across heterogeneous clinical datasets.

Despite recent progress, however, EEG FMs remain predominantly centered around masked reconstruction losses, with comparatively limited exploration of contrastive and hybrid objectives at scale \cite{Kuruppu2025-nk, Weng2025-an}. Thus, \textit{CoCoT-EEG is so far the only EEG FM employing multi-scale convolutions at the signal level across patches, and is further improved with contrastive pretraining}.

\subsection{Limitations and future work}

Several limitations warrant discussion: First, our pretraining datasets (TUEG: ~27,000 hours; HBN: ~1,500 hours) are smaller than those used by some recent EEG foundation models, leaving open questions about how contrastive objectives scale with data. Second, our augmentation ablations were conducted at reduced scale; the relative importance of augmentations may shift with larger models and datasets.
Third, while our direct comparison of the MoCo vs. MAE objective provides strong evidence in favor of contrastive pretraining in general, there may be tasks where reconstruction or hybrid objectives may be more appropriate, e.g., with high-SNR signals like epileptic events.
Despite these limitations, EEG FMs hold substantial promise for clinical and societal impact: by learning transferable representations from large unlabeled corpora, they can lower the data requirements for downstream applications, potentially expanding access to neurological screening in settings where expert-labeled data and trained clinicians are scarce.

A natural next step is therefore scaling contrastive pretraining to substantially larger and more heterogeneous corpora that span clinical EEG (e.g., TUEG), developmental cohorts (e.g., HBN), sleep recordings (e.g., MASS, SHHS), and other BCI datasets. To do so robustly raises several concrete challenges: harmonizing across heterogeneous montages and channel counts (from 4-channel consumer devices to 256-channel research caps), reconciling sampling rates and reference schemes, accounting for site- and device-specific noise distributions, and designing batch-construction strategies that prevent dominant data sources from collapsing the contrastive objective.

\clearpage
\begingroup
\small

\bibliographystyle{unsrt}
\bibliography{paperpile.bib,new_refs}
\endgroup

\newpage
\appendix

\section{Pretraining hyperparameters}\label{app:pretrainHP}
\begin{table}[h]
\caption{Pretraining hyperparameters for TUEG and HBN.}
\label{tab:pretrain-hparams}
\vskip 0.15in
\begin{center}
\begin{small}
\begin{sc}
\begin{tabular}{lcc}
\toprule
Hyperparameter & TUEG & HBN \\
\midrule
\multicolumn{3}{l}{\textbf{\textit{Optimization}}} \\
Learning rate & $1 \times 10^{-4}$ & $5 \times 10^{-5}$ \\
Batch size & 1336 & 800 \\
Weight decay & $1 \times 10^{-4}$ & $1 \times 10^{-4}$ \\
Warmup steps & 3000 & 1500 \\
Max steps & 95k & 95k \\
\midrule
\multicolumn{3}{l}{\textbf{\textit{MoCo Loss}}} \\
Temperature & 0.07 & 0.07 \\
Queue size & 65536 & 65536 \\
Feature dimension & 128 & 128 \\
\midrule
\multicolumn{3}{l}{\textit{\textbf{CoCoT-EEG}}} \\
Patch size & 50 & 50 \\
Model dimension & 512 & 512 \\
FFN dimension & 2048 & 2048 \\
Layers & 12 & 12 \\
Attention heads & 8 & 8 \\
Drop path rate & 0.1 & 0.1 \\
\midrule
\multicolumn{3}{l}{\textbf{\textit{Projector}}} \\
Hidden dimension & 2048 & 2048 \\
Output dimension & 128 & 128 \\
Layers & 2 & 2 \\
\midrule
\multicolumn{3}{l}{\textbf{\textit{Data}}} \\
Sampling rate & 100 Hz & 100 Hz \\
Window size & 5 seconds & 5 seconds \\
Window stride & 5 seconds & 5 seconds \\
\bottomrule
\end{tabular}
\end{sc}
\end{small}
\end{center}
\vskip -0.1in
\end{table}

\section{Finetuning hyperparameters}\label{app:finetuneHP}
\begin{table}[h]
\caption{Finetuning hyperparameters for TUEG and HBN.}
\label{tab:pretrain-hparams}
\vskip 0.15in
\begin{center}
\begin{small}
\begin{sc}
\begin{tabular}{lc}
\toprule
Hyperparameter & Value \\
\midrule
\multicolumn{2}{l}{\textbf{\textit{Optimization}}} \\
Learning rate & $1 \times 10^{-4}$ (except CHB-MIT = 1e-7) \\
Batch size & 16 \\
Weight decay & $5 \times 10^{-2}$ \\
\midrule
\multicolumn{2}{l}{\textbf{\textit{Classification head}}} \\
Hidden dimension & 1024 \\
Layers & 2 \\
\midrule
\multicolumn{2}{l}{\textbf{\textit{Data}}} \\
Sampling rate & 100 Hz \\
Sequence length & Depends on the dataset \\
\bottomrule
\end{tabular}
\end{sc}
\end{small}
\end{center}
\vskip -0.1in
\end{table}

\section{Computing resources}\label{app:computing_resources}
Pretraining on either TUEG and HBN dataset was done using a H100-SXM-8-80G GPU machine for 95,000 steps. It took around 15 hours. In total, counting all the different ablations ran, around 20 pretrainings had to be run, amounting for 300 hours of H100-SXM-8-80G.

Finetuning on all the downstream datasets was done using a H100-SXM-8-80G GPU machine would take around 4 hours per checkpoint. Accounting for all the ablations ran in the paper, around 100 hours of H100-SXM-8-80G was used for finetuning. 

In total, approximately 400 hours of H100-SXM-8-80G were used for the experimental results of this paper.

\section{EEG-specific augmentations}
\label{app:augmentations}

To construct positive pairs for contrastive pretraining, we apply random compositions of augmentations designed for EEG signals. These transformations either mimic common acquisition artifacts and noise sources encountered in EEG recordings, or encourage invariance to nuisance variability while preserving task-relevant neural activity.

\paragraph{Channel dropout.}
A random subset of channels is zeroed independently with probability $p_{\mathrm{drop}}=0.15$. This augmentation simulates corrupted or disconnected electrodes and encourages robustness to missing sensors.

\paragraph{Gaussian noise.}
Independent Gaussian noise with standard deviation $\sigma=0.08$ is added to the input signal, modeling acquisition and environmental noise commonly present in EEG recordings.

\paragraph{Time masking.}
One or two contiguous temporal spans are masked by setting the signal to zero. Mask lengths are sampled uniformly between $8\%$ and $15\%$ of the sequence length. This augmentation encourages robustness to transient artifacts and partial temporal occlusions.

\paragraph{Amplitude scaling and offset.}
Each channel undergoes random amplitude scaling and additive offset perturbations. Scaling factors are sampled uniformly in $[0.85, 1.15]$, while offsets are sampled proportionally to the channel standard deviation. This models variability in electrode impedance and recording conditions.

\paragraph{Band-stop filtering.}
A random narrow frequency band is attenuated in the Fourier domain using a smooth band-stop filter. The stop-band center is sampled uniformly between 4 and 40 Hz, with widths between 2 and 6 Hz. This augmentation encourages robustness to frequency-specific corruption and reduced dependence on isolated spectral components.

\paragraph{Phase jitter.}
Small random perturbations are applied to the STFT phase while preserving spectral magnitudes. Phase offsets are sampled uniformly between $3^\circ$ and $8^\circ$. This introduces controlled spectral distortions while approximately preserving the signal power spectrum.

We study the contribution of these augmentations through leave-one-out and leave-one-in ablations in Sec.~\ref{sec:results_augs}.

\section{MAE Pretraining} \label{app:mae}

For reconstruction-based pretraining, we adopt the masked EEG reconstruction strategy and preprocessing pipeline of CBraMod \cite{Wang2024-sh} with minimal modifications. EEG recordings are segmented into fixed-length temporal patches, a random subset of patches is masked, and the encoder is trained to reconstruct only the masked patches using a mean squared error objective. Following CBraMod, reconstruction is performed through a lightweight linear reconstruction head applied to the learned patch representations. We use the same masking ratio, normalization strategy, and preprocessing procedure as CBraMod.

Recordings are band-pass filtered (0.3--75 Hz), notch filtered at 60 Hz, resampled (at 100Hz in our setting), segmented into fixed windows, and samples containing extreme amplitudes are removed through an automated artifact rejection procedure. EEG amplitudes are then divided by $100$ to approximately lie in the range $[-1,1]$. As noted by CBraMod, reconstruction-based pretraining on TUEG is particularly sensitive to noisy or artifact-heavy recordings, which can substantially degrade downstream performance if not filtered from the pretraining corpus.

\section{3D Spatial Positional Encoding}
\label{app:spatialPE}

To evaluate whether explicit electrode topology improves representation learning, we augment CoCoT-EEG with a 3D spatial positional encoding (3DPE) inspired by REVE \cite{El_Ouahidi_undated-mu}. Unlike REVE's joint 4D spatiotemporal encoding, we encode only spatial coordinates; temporal patch positions are handled separately using a temporal Fourier positional encoding.

Each EEG channel \(c\) is associated with a standardized 3D electrode coordinate
\[
\mathbf{r}_c = (x_c, y_c, z_c) \in \mathbb{R}^3.
\]
For every coordinate, we compute a Fourier embedding over the Cartesian product of spatial frequencies:
\[
\gamma(\mathbf{r}_c)
=
\left[
\cos(\theta),
\sin(\theta)
\right],
\]
where the phase \(\theta\) is constructed from frequency combinations over \((x_c,y_c,z_c)\), following the spatial component of REVE. We use \(F=4\) frequencies per dimension, resulting in \(2F^3=128\) Fourier features per channel. When this dimensionality differs from the transformer dimension, the Fourier embedding is projected with a bias-free linear layer.

Following REVE, the Fourier representation is combined with a learnable projection of the raw coordinates:
\[
\mathbf{p}^{\mathrm{spatial}}_c
=
\mathrm{LayerNorm}
\left(
W_\gamma \gamma(\mathbf{r}_c)
+
\mathrm{MLP}(\mathbf{r}_c)
\right),
\]
where \(\mathrm{MLP}\) is implemented as a bias-free linear layer followed by GELU and LayerNorm. During training, we also add small Gaussian noise to the coordinates before encoding.

After multiscale convolution and temporal patchification, the spatial embedding is added to each token together with the temporal Fourier positional encoding:
\[
\tilde{\mathbf{h}}_{b,c,s}
=
\mathbf{h}_{b,c,s}
+
\mathbf{p}^{\mathrm{spatial}}_c
+
\mathbf{p}^{\mathrm{temporal}}_s.
\]

To test whether the model truly exploits meaningful electrode topology rather than simply relying on additional channel-identifying embeddings, we evaluate a shuffled-position control setting. In this condition, electrode coordinates are randomly permuted once at finetuning initialization while preserving the original EEG channel order. This destroys the anatomical correspondence between electrodes and spatial coordinates while retaining individual channel identifiers.

\section{Downstream task datasets}\label{app:downstream}
The list of the different downstream datasets can be found in Table \ref{tab:downstream-tasks}.

\begin{table}
\caption{Overview of downstream task datasets.}
\label{tab:downstream-tasks}
\begin{center}
\resizebox{0.70\columnwidth}{!}{
\begin{tiny}
\begin{sc}
\begin{tabular}{@{}lccccc@{}}
\toprule
Dataset & Task & \#Ch. & Dur. & \#Samp. & \#Cls. \\
\midrule
BCI-IV & Motor Im. & 22 & 8s & 20K & 4\\
PHYSIONET & Motor Im. & 64 & 4s & 10K & 4 \\
SHU & Motor Im. & 32 & 4s & 12K & 2 \\
FACED & Emotion & 32 & 10s & 10K & 9 \\
SEED-V & Emotion & 62 & 1s & 118K & 5 \\
CHB-MIT & Seizure & 16 & 10s & 327K & 2 \\
TUAB & Abnormal & 16 & 10s & 409K & 2 \\
SEED-VIG & Vigilance & 17 & 8s & 20K & Regr. \\
HMC & Sleep Staging & 4 & 30s & 137K & 4 \\
ISRUC & Sleep Staging & 6 & 30s & 90K & 5 \\
\bottomrule
\end{tabular}
\end{sc}
\end{tiny}
}
\end{center}
\vskip -0.1in
\end{table}

\paragraph{Motor Imagery Classification.}
We evaluate on three motor imagery datasets: \texttt{BCI-IV} (BCI Competition IV-2a)~\cite{tangermann2012review}, \texttt{PHYSIO} (PhysioNet Motor Movement/Imagery)~\cite{Goldberger2000-ao}, and \texttt{SHU-MI}~\cite{Ma2022-hr}. BCI Competition IV-2a contains 22-channel EEG from 9 subjects performing four motor imagery tasks (left hand, right hand, feet, tongue). PhysioNet-MI comprises 64-channel recordings from 109 subjects across four classes (left fist, right fist, both fists, both feet). SHU-MI provides 32-channel binary classification data (left/right hand) from 25 subjects. All signals are resampled to 100 Hz and segmented into fixed-duration trials.

\paragraph{Emotion Classification.}
We use \texttt{FACED}~\cite{chen2023large} and \texttt{SEED-V}~\cite{liu2021comparing} for emotion recognition evaluation. FACED is a large-scale affective computing dataset with 32-channel EEG from 123 subjects, covering nine fine-grained emotion categories (amusement, inspiration, joy, tenderness, anger, fear, disgust, sadness, neutral). SEED-V contains 62-channel EEG from 16 subjects across three sessions, with five emotion categories (happy, sad, neutral, disgust, fear).

\paragraph{Seizure Detection.}
We employ \texttt{CHB-MIT} (CHB-MIT Scalp EEG Database)~\cite{shoeb2009application} for seizure detection evaluation. This clinical dataset contains 16-channel EEG recordings from pediatric patients with intractable seizures, providing a challenging binary classification task with highly imbalanced class distributions.

\paragraph{Abnormality Detection.}
\texttt{TUAB} is a part of the Temple University database \cite{Obeid2016-az}, where 16-channel EEG segments are labeled as pathological or non-pathological, similar to seizure detection but in a more general medical context.

\paragraph{Sleep staging.}
We use \texttt{HMC}~\cite{Alvarez-Estevez2021-df} and \texttt{ISRUC}~\cite{Khalighi2016-lr} for sleep stage classification evaluation. HMC contains 151 whole-night polysomnographic (PSG) recordings collected from a heterogeneous population referred for clinical PSG examination. ISRUC consists of PSG recordings from 118 adults across three subgroups---patients with various sleep disorders, longitudinal recordings, providing complementary distributional coverage to HMC.

\paragraph{Vigilance Estimation.}
The \texttt{SEED-VIG} dataset~\cite{Zheng2017-tu} provides continuous vigilance labels derived from PERCLOS (percentage of eye closure) during simulated driving. This 17-channel dataset enables regression-based evaluation of drowsiness estimation, collected from 23 subjects over approximately 2-hour driving sessions.

\clearpage
\section{From scratch Training Ablation of Truncated Architectures}
\label{app:trunc_scratch}
\begin{table}[H]
\centering
\caption{
Scratch training comparison between truncated Conv-first and Patch-first architectures.
Balanced accuracy (\%) is reported for all tasks except SEED-VIG (Pearson correlation $\times 100$).
Values are mean $\pm$ 95\% CI. Bold denotes the best result per row.
}
\label{tab:trunc_scratch}

\small
\setlength{\tabcolsep}{5pt}

\begin{tabular}{lcc}
\toprule
Dataset & Patch-first Scratch & Conv-first Scratch \\
\midrule
BCI-IV   & 46.83 {\tiny $\pm$ 2.34} & \textbf{48.09} {\tiny $\pm$ 2.88} \\
FACED    & 50.39 {\tiny $\pm$ 0.95} & \textbf{53.75} {\tiny $\pm$ 0.76} \\
PHYSIO   & 60.15 {\tiny $\pm$ 0.48} & \textbf{61.23} {\tiny $\pm$ 0.71} \\
SEED-V   & 39.92 {\tiny $\pm$ 0.94} & \textbf{40.45} {\tiny $\pm$ 0.16} \\
CHB-MIT  & 51.57 {\tiny $\pm$ 4.99} & \textbf{53.77} {\tiny $\pm$ 5.62} \\
SHU      & \textbf{61.46} {\tiny $\pm$ 2.27} & 60.84 {\tiny $\pm$ 1.07} \\
SEED-VIG & \textbf{63.75} {\tiny $\pm$ 3.56} & 57.40 {\tiny $\pm$ 5.23} \\
\bottomrule
\end{tabular}
\end{table}

\section{From scratch, single-task results}\label{app:fromscratch}
\begin{table*}[h]
\caption{Performance comparison including additional baselines from secondary sources. Balanced accuracy is displayed in \% (mean ± 95\% CI where available). Baseline performance is highlighted in light gray.}
\label{tab:expanded_from_scratch_baselines}
\vskip 0.15in
\begin{center}
\begin{small}
\begin{sc}
\setlength{\aboverulesep}{0pt}
\setlength{\belowrulesep}{0pt}
\newcolumntype{g}{>{\columncolor[gray]{0.9}}c}
\begin{tabular}{lggggc}
\toprule
\rowcolor{white}
 & \multicolumn{4}{g}{Baselines} & CoCoT-EEG \\
\cmidrule{2-5} \cmidrule{6-6}
\rowcolor{white}
Dataset & \cellcolor[gray]{0.9}ContraWR & \cellcolor[gray]{0.9}CNN-Trans & \cellcolor[gray]{0.9}ST-Trans & \cellcolor[gray]{0.9}SimpleConv & From Scratch \\
\midrule
BCIIV   & 46.78 & 46.00 & 45.75 & \textbf{62.38\text{*} {\scriptsize $\pm$ 2.38}} & \underline{54.40} {\scriptsize $\pm$ 2.99} \\
CHB-MIT & \underline{63.44} & \textbf{63.89} & 59.15 & 59.78\text{*} {\scriptsize $\pm$ 2.17} & 60.55 {\scriptsize $\pm$ 0.79} \\
FACED   & \underline{48.87} & 46.97 & 48.10 & 39.51\text{*} {\scriptsize $\pm$ 2.84} & \textbf{55.56 {\scriptsize $\pm$ 0.73}} \\
MUMTAZ  & 91.95 & \underline{93.05} & 91.35 & 56.09\text{*} {\scriptsize $\pm$ 1.96} & \textbf{93.56 {\scriptsize $\pm$ 0.44}} \\
PHYSIO  & 58.92 & \underline{60.53} & 60.35 & 60.29\text{*} {\scriptsize $\pm$ 0.62} & \textbf{62.44 {\scriptsize $\pm$ 1.31}} \\
SEED-V  & 35.46 & \underline{36.78} & 30.52 & 19.97\text{*} {\scriptsize $\pm$ 0.04} & \textbf{41.41 {\scriptsize $\pm$ 0.44}} \\
SHU     & 58.73 & 59.75 & 59.92 & \textbf{62.92\text{*} {\scriptsize $\pm$ 1.33}} & 62.25 {\scriptsize $\pm$ 1.65} \\
TUAB    & 77.46 & 77.77 & 79.66 & \textbf{81.74\text{*} {\scriptsize $\pm$ 0.39}} & \underline{79.17} {\scriptsize $\pm$ 0.14} \\
SEED-VIG (Corr) & 52.35 & 58.29 & \underline{60.20} & --- & \textbf{60.37 {\scriptsize $\pm$ 2.04}} \\
\bottomrule
\end{tabular}
\end{sc}
\end{small}
\end{center}
\vskip -0.1in
\end{table*}

\section{Pretraining dataset ablation results}
\begin{table*}[h]
\label{tab:fullwithhbn}
\caption{Performance of CoCoT-EEG trained from scratch, or pretrained (PT) with MoCo on TUEG or HBN and finetuned on each dataset. Balanced accuracy (in \%, mean ± 95\% CI) is reported for all tasks except SEED-VIG (Pearson correlation).
LaBraM \cite{Jiang2024-qb}, CBraMod: mean ± 95\% CI \cite{Wang2024-sh}, REVE: mean ± s.t.d. \cite{Ouahidi2025-gf}, CSBrain: mean ± s.t.d. \cite{Zhou2025-hl}. \textbf{Bold} denotes SOTA mean. FT: end-to-end fine-tuning.}
\label{tab:main_results_extended}
\vskip 0.15in
\begin{center}
\begin{small}
\begin{sc}
\setlength{\aboverulesep}{0pt}
\setlength{\belowrulesep}{0pt}
\setlength{\tabcolsep}{2.8pt}
\newcolumntype{g}{>{\columncolor[gray]{0.9}}c}
\begin{tabular}{lggggclcc}
\toprule
 & \multicolumn{4}{g}{Baselines} & & \multicolumn{3}{c}{CoCoT-EEG (OURS)} \\
\cmidrule{2-5} \cmidrule{7-9}
\rowcolor{white}
Dataset & \cellcolor[gray]{0.9}\small LaBraM & \cellcolor[gray]{0.9}\small CBraMod & \cellcolor[gray]{0.9}\small REVE & \cellcolor[gray]{0.9}\small CSBrain & & \small Scratch & \small TUEG PT & \small HBN PT \\
\midrule
BCI-IV   & 47.48 {\tiny $\pm$ 0.85} & 51.38 {\tiny $\pm$ 0.66} & \textbf{63.96} {\tiny $\pm$ 0.95} & 56.57 {\tiny $\pm$ 0.71} & & 54.40 {\tiny $\pm$ 2.99} & 60.55 {\tiny $\pm$ 1.40} & 56.08 {\tiny $\pm$ 3.01} \\
CHB-MIT  & 70.75 {\tiny $\pm$ 3.58} & 73.98 {\tiny $\pm$ 2.84} & --- & 72.62 {\tiny $\pm$ 1.15} & & 60.55 {\tiny $\pm$ 0.79} & \textbf{82.35} {\tiny $\pm$ 0.61} & 80.95 {\tiny $\pm$ 0.60} \\
FACED    & 52.73 {\tiny $\pm$ 1.07} & 55.09 {\tiny $\pm$ 0.89} & 56.46 {\tiny $\pm$ 1.64} & 57.52 {\tiny $\pm$ 0.42} & & 55.56 {\tiny $\pm$ 0.73} & \textbf{57.73} {\tiny $\pm$ 2.04} & 56.15 {\tiny $\pm$ 1.20} \\
HMC      & 72.77 {\tiny $\pm$ 1.01} & 72.69 {\tiny $\pm$ 0.41} & --- & 73.45 {\tiny $\pm$ 0.47} & & 73.69 {\tiny $\pm$ 0.21} & \textbf{76.13} {\tiny $\pm$ 0.11} & 75.79 {\tiny $\pm$ 0.21} \\
ISRUC    & 76.33 {\tiny $\pm$ 1.02} & 78.65 {\tiny $\pm$ 1.1} & --- & 79.25 {\tiny $\pm$ 0.30} & & 74.62 {\tiny $\pm$ 0.32} & \textbf{79.97} {\tiny $\pm$ 0.34} & 79.20 {\tiny $\pm$ 0.15} \\
PHYSIO   & 61.73 {\tiny $\pm$ 1.22} & 64.17 {\tiny $\pm$ 0.91} & 64.80 {\tiny $\pm$ 1.50} & 63.04 {\tiny $\pm$ 0.90} & & 62.44 {\tiny $\pm$ 1.31} & \textbf{65.31} {\tiny $\pm$ 1.42} & 64.64 {\tiny $\pm$ 0.73} \\
SEED-V   & 39.76 {\tiny $\pm$ 1.38} & 40.91 {\tiny $\pm$ 0.97} & --- & 41.97 {\tiny $\pm$ 0.33} & & 41.41 {\tiny $\pm$ 0.44} & \textbf{42.13} {\tiny $\pm$ 0.36} & 40.74 {\tiny $\pm$ 0.62} \\
SHU      & 61.66 {\tiny $\pm$ 1.92} & 63.70 {\tiny $\pm$ 1.51} & --- & 64.17 {\tiny $\pm$ 0.37} & & 62.25 {\tiny $\pm$ 1.65} & \textbf{64.74} {\tiny $\pm$ 0.87} & 64.07 {\tiny $\pm$ 0.50} \\
TUAB     & 81.40 {\tiny $\pm$ 0.19} & 82.49 {\tiny $\pm$ 0.25} & \textbf{83.15} {\tiny $\pm$ 0.15} & 81.72 {\tiny $\pm$ 0.43} & & 79.17 {\tiny $\pm$ 0.14} & 82.78 {\tiny $\pm$ 0.25} & 81.37 {\tiny $\pm$ 0.14} \\
SEED-VIG & 63.47 {\tiny $\pm$ 1.35} & 55.02 {\tiny $\pm$ 1.15} & --- & 63.14 {\tiny $\pm$ 3.56} & & 60.37 {\tiny $\pm$ 2.04} & \textbf{64.43} {\tiny $\pm$ 1.12} & 61.66 {\tiny $\pm$ 2.53} \\
\bottomrule
\end{tabular}
\end{sc}
\end{small}
\end{center}
\vskip -0.1in
\end{table*}

\end{document}